\pdfoutput=1

\documentclass[11pt]{article}

\usepackage[final]{emnlp2021}

\usepackage{times}
\usepackage{latexsym}
\usepackage{amsmath}
\usepackage[T1]{fontenc}
\usepackage{graphicx}

\usepackage{caption}
\usepackage{subcaption}

\usepackage[utf8]{inputenc}
\usepackage{microtype}
\author{Sunipa Dev \\
  she/her \\
  UCLA \\
  \And
  Masoud Monajatipoor*\\
  he/him \\
UCLA\\

\And Anaelia Ovalle*\\
they/he/she\\
UCLA\\

\And Arjun Subramonian*\\
they/them\\
UCLA, Queer in AI\\

\AND Jeff M Phillips\\
he/him\\
University of Utah\\

\And Kai-Wei Chang\\
he/him\\
UCLA\\

}

\usepackage{diagbox}
\usepackage{adjustbox}
\usepackage{enumitem}
\usepackage{xcolor}

\title{Harms of Gender Exclusivity and Challenges in Non-Binary Representation in Language Technologies}

\date{}

\begin{document}

\maketitle

\begin{abstract}
{\color{red} \emph{Content Warning:} This paper contains examples of stereotypes and associations, misgendering, erasure, and other harms that could be offensive and triggering to trans and non-binary individuals.}

Gender is widely discussed in the context of language tasks and when examining the stereotypes propagated by language models. However, current discussions primarily treat gender as binary, which can perpetuate harms such as the cyclical erasure of non-binary gender identities. These harms are driven by model and dataset biases, which are consequences of the non-recognition and lack of understanding of non-binary genders in society.
In this paper, we explain the complexity of gender and language around it, and survey non-binary persons to understand harms associated with the   
treatment of gender as binary in English language technologies. We also detail how current language representations (e.g., GloVe, BERT) capture and perpetuate these harms and related challenges that need to be acknowledged and addressed for representations to equitably encode gender information. 
\end{abstract}

\section{Introduction}

\let\thefootnote\relax\footnotetext{* Equal contribution}

\let\thefootnote\relax\footnotetext{\{sunipa,anaelia,arjunsub,kwchang\}@cs.ucla.edu,\\
 monajati@g.ucla.edu,
jeffp@cs.utah.edu
}

As language models are more prolifically used in language processing  applications, ensuring a higher degree of fairness in associations made by their learned representations and intervening in any biased decisions they make has become increasingly important. Recent work analyzes, quantifies, and mitigates language model biases such as gender, race or religion-related stereotypes in static word embeddings (GloVe~\cite{pennington2014glove}) and contextual (e.g., BERT~\cite{devlin-etal-2019-bert}) representations~\cite{bolukbasi2016man,De-artega2019bias,ravfogel2020null,dev2020oscar}. 

A bulk of social bias studies on language models have focused on binary gender and the stereotypes associated with masculine and feminine attributes~\cite{bolukbasi2016man,gap,dev2020oscar}. Additionally, models often rely on gendered information for decision making, such as in named entity recognition, coreference resolution, and machine translation \cite{10.1145/3372923.3404804, zhao-etal-2018-gender, stanovsky-etal-2019-evaluating}, but the purview of gender in these tasks and associated measures of performance focus on binary gender. While discussing binary gender bias and improving model performance are important, it is important to reshape our understanding of gender in language technologies in a more accurate, inclusive, non-binary manner. 

Current language models can perpetrate harms such as the cyclical erasure of non-binary gender identities \cite{uppunda2021adapting,sap2021, alma991057948793106011, fiske1993controlling, fast2016shirtless, BehmMorawitz2008MeanGT}. These harms are driven by model and dataset biases due to tainted examples, limited features, and sample size disparities \citep{wang2019balanced, barocas-hardt-narayanan, intersectional}, which are consequences of the non-recognition and a lack of understanding of non-binary genders in society \citep{criminalization, 10.7312/raju18532}. 

Some recent works attempt to mitigate these harms by building task-specific datasets that are not restricted to binary gender and building metrics that on extension, could potentially measure biases against all genders \cite{cao2020towards,rudinger2018gender}. While such works that intentionally inject real-world or artificially-created data of non-binary people into binary-gendered datasets are well-intentioned, 
they 
could benefit from a broader perspective of harms as perceived by non-binary persons to avoid
mischaracterizing non-binary genders as a single gender~\cite{sun2021they} or perpetuating biases through non-intersectional training examples, i.e. examples that do not capture the interconnected nature of social identities \cite{897196}.

In this paper, we conduct a detailed investigation into the representational and allocational harms \citep{barocas2017problem, blodgett2020language} related to the treatment of gender with binary predilections in English language technologies. We do so by explaining the complexity of gender and language around it, and surveying non-binary persons with some familiarity with AI on potential harms in common NLP tasks.  
While the challenges associated with limited or tainted data are loosely hypothesized, they are not well understood. 

We study the extent of these data challenges and detail how they manifest in the resultant language representations and downstream tasks. 
We examine both static embeddings (GloVe) and contextual representations (BERT) with respect to the quality of representations (Section \ref{sec: representation}) of non-binary-associated words and pronouns.
 We highlight how the disparity in representations cyclically propagates the biases of underrepresentation and misrepresentation and can lead to the active misgendering and erasure of non-binary persons in language technologies.


\section{Gender, Language, and Bias}
We first discuss the complex concepts of gender and bias and their expression in English language.

\subsection{Gender}
\label{sec:gender}
In this paper, \emph{gender} refers to \emph{gender identity}, as opposed to \emph{gender expression} or \emph{sex}. 
\emph{Gender identity} concerns how individuals experience their own gender. In contrast, \emph{gender expression} concerns how one expresses themselves, through their ``hair length, clothing, mannerisms, makeup'' and \emph{sex} relates to one's  ``genitals, reproductive organs, chromosomes, hormones, and secondary sex  characteristics'' \citep{10.7312/raju18532}.
Gender identity, gender expression, and sex do not always ``align'' in accordance with Western cisnormativity \citep{10.7312/raju18532}. 
However, 
people are conditioned to erroneously believe otherwise,
which leads to ``societal expectations and stereotypes around gender roles'' and the  compulsive \emph{(mis)gendering} of others \citep{cao2020towards, serano2007}.

\paragraph{Gender in Western Society}

In Western society, \emph{discourse} around one's gender identity can, but does not always, comprise two intersecting aspects:
    
    \noindent (i) a description of how it is similar to or different from the binary genders, i.e. male and female. For instance, genderfluid persons do not identify with a single gender, and agender individuals do not subscribe to gender at all \citep{10.7312/raju18532}. It is important to note that gender may fluctuate over an individual's lifetime, and it is extremely problematic to assume a biologically essentialist view of it \citep{weber2019}, and
    \noindent (ii)  whether it is the same as or differs from the individual's gender assigned at birth, i.e. cisgender or transgender, respectively. Many individuals who are not cis, including non-binary people, identify as trans.
 
Non-binary genders encompass \emph{all} the genders that do not conform to the Western gender binary \citep{10.7312/raju18532}. There are many non-Western non-cis identities, like the Jogappas of Karnataka, Muxes of Oaxaca, and Mahuwahines of Hawai'i \citep{jogappa, muxes, clarke2019}. However, non-Western non-cis identities cannot be accurately described by the Western-centric, English-based gender framework afore established \citep{muxes, thorne2019}. Hence, as this paper focuses on the English language, its treatment of non-binary genders does \emph{not} adequately include non-Western non-cis identities.

\paragraph{Pronouns and Gendered Names}

In societies where language has referential gender, i.e., when an entity is referred to, and ``their gender (or sex) is realized linguistically''~\cite{cao2020towards}, it is difficult to escape gendering others. In English, pronouns are gendered; hence, pronouns can be central to English speakers' gender identity.  
However, pronouns cannot be bijectively mapped to gender. For example, not all non-binary persons use $they/them/theirs$ pronouns, nor do all persons who use $they/them/theirs$ pronouns identify as non-binary \citep{clarke2019}. Furthermore, the use of binary pronouns, $he$ and $she$, is not exclusive to cis individuals; trans and non-binary individuals also use them. English pronouns are always evolving \citep{pronouns2016}. Singular $they$ has become widely adopted by trans and non-binary persons \citep{pronouns2016, feraday2016, clarke2019}. Neopronouns like $xe/xem/xyr$ and $ze/hir/hirs$ are also in use by non-cis individuals \citep{feraday2016}.

Not everyone who speaks English chooses to use pronouns, and some individuals use multiple sets of pronouns (e.g. $she/her/hers$ and $they/them/theirs$) \citep{feraday2016}. Many non-binary people use different pronouns depending on the space in which they are, especially if they are not publicly out; for example, a non-binary person may accept $she/her$ pronouns at work but use $they/them$ pronouns outside of work. Additionally, non-binary people can find multiple sets of pronouns affirming; for instance, non-binary men may use a combination of $they/them/theirs$ and $he/him/his$. Furthermore, genderfluid individuals can use different sets of pronouns based on their ``genderfeels'' at a certain time \cite{vasundhara2021}. This may also lead individuals to be open to being referenced by ``all pronouns'' or ``any pronouns.'' Ultimately, individuals use the pronouns that allow them to feel gender euphoria in a given space, at a given time \cite{vasundhara2021}.

In languages without referential gender or where pronouns are seldom used (e.g. Estonian), pronouns can be less central to one's gender identity \citep{estonian}. 

Another form of referential gender is gendered names, which are assumed for binary gender, even in language technologies, which 
 itself can be inaccurate and problematic. Additionally, trans and non-binary persons may choose a new name that matches their gender identity to replace their \emph{deadname}, i.e. name assigned at birth  \citep{enbynames}. Many Western non-binary chosen names are creative and diverse, overlapping with common nouns or nature words, having uncommon orthographic forms, and/or consisting of a single letter  \citep{enbynames}.

\paragraph{Lexical Gender}
Lexical gender in English language is gender (or sex) conveyed in a non-referential manner~\cite{cao2020towards}. Examples include ``mother'' and ``Mr.'' Non-binary persons have adopted honorifics like ``Mx.'' to eliminate gendering \citep{clarke2019}, and often use gender-neutral terms like ``partner'' to refer to their significant other. However, their adoption into written text and narratives is recent and sparse. 

\paragraph{Implications in Language Technologies}
Given the complex and evolving nature of gender and the language around it, for language technologies to truly equitably encode gender, they would need to capture the full diversity and flexibility therein. 



\subsection{Biases}
There has been an increase in awareness of the social biases that language models carry. 
In this paper, we use the term \emph{bias} to refer to a skewed and undesirable association in language representations which has the potential to cause representational or allocational harms \citep{barocas2017problem}.
There have been multiple attempts to understand social biases in language processing  \cite{sheng2021societal,Caliskan183}, quantify them~\cite{rudinger2018gender, gap, De-artega2019bias}, and mitigate them~\cite{zhao2019gender, ravfogel2020null,sun2019mitigating}. A primary focus has been on gender bias, but the narrative has been dominated by biases associated with binary gender, primarily related to occupations and adjectives. However, 
the biases faced by non-binary persons can be distinct from this.
Non-binary genders are severely underrepresented in textual data, which causes language models to learn meaningless, unstable representations for non-binary-associated pronouns and terms. Furthermore, there are derogatory adjectives associated with non-binary-related terms (as seen in Appendix \ref{app: representational skews}). Thus, analyzing and quantifying biases associated with non-binary genders cannot be treated merely as a corollary of those associated with binary gender. 


 
\section{Harms}  
Utilizing and perpetuating the binary construction of gender in English in language technologies can have adverse impacts. We focus on specific tasks within language processing and associated applications in human-centered domains where harms can be perpetrated, motivated by their frequent mention in a survey we conduct (Section \ref{sec:survey}).  
The primary harms we discuss are misgendering and erasure.

\noindent \paragraph{\textbf{Misgendering:}} Misgendering is the act of accidentally or intentionally addressing someone (oneself or others) using a gendered term that does not match their gender identity. 
Misgendering persons and the associated harms have been studied in contexts of computer vision~\cite{keyes2018misgendering} and human-computer interaction~\cite{keyes2021truth}, which highlight its adverse impact on the mental health of non-binary individuals.
Language applications and their creators can also perpetrate misgendering. For instance, language applications that operationally ask non-binary users to choose between $male$ and $female$ as input force non-binary users to misgender themselves \citep{keyes2018misgendering, katta2019}. Furthermore,  language models which do not explicitly collect gender information are capable of both accidental and intentional misgendering. Specifically, language models accidentally misgender non-binary persons when there is insufficient information to disambiguate the gender of an individual, and so they default to binary pronouns and binary-gendered terms, potentially based on stereotypes. 
However, as shown in Section \ref{sec: representation}, language models can also misgender non-binary individuals even when their pronouns are provided.

\noindent \paragraph{\textbf{Erasure:}} In one sense, erasure is the accidental or intentional invalidation or obscuring of non-binary gender identities.
For example, the language technology Genderify, which purportedly ``identif[ied] someone’s [binary] gender based on their name, email address or username'' erased non-binary people by reductively distributing individuals into binary ``gender bins'' by their name, based on the assumption that they were cisgender \citep{lauer2020, katta2019, serano2007}. Another sense of erasure is in how stereotypes about non-binary communities are portrayed and propagated (see Appendix Table \ref{App: derogatory}). Since non-binary individuals are often ``denied access to media and economic and political power,''  individuals in power can paint negative narratives of non-binary persons or erase the diversity in gender communities \citep{serano2007, 10.7312/raju18532}. 

Language applications are capable of automating erasure, in a cyclical fashion \cite{pmlr-v80-hashimoto18a, sap2021}. We posit the cycle of non-binary erasure in text, in which:
\noindent (i) language applications, trained on large, binary-gendered corpora, reflect the misgendering and erasure of non-binary communities in real life \citep{alma991057948793106011, fiske1993controlling}
\noindent (ii) this reflection is viewed as a ``source of truth and scientific knowledge''~\citep{keyes2021truth} 
\noindent (iii) consequently, authors buy into these harmful ideas and other language models encode them, leading them to stereotypically portray non-binary characters in their works or not include them at all, and \citep{fast2016shirtless}
\noindent (iv) this further amplifies non-binary erasure, and the cycle continues.

\subsection{Survey on Harms}
\label{sec:survey}

To understand harms associated with skewed treatment of gender in English NLP tasks and applications, the perspective of those facing the harms is essential. We conduct a survey for the same.

\paragraph{Survey Respondents}
We focused this survey on non-binary persons who have familiarity with AI. We acknowledge that this indeed is a limitation, as it narrows our focus to non-binary persons of specific socioeconomic status, ethnicity, and English fluency. However, we field  this survey as the first in a series to gain foray into harms experienced by non-binary individuals who build AI and know its effects. Furthermore, it allows us to gather what tasks could potentially cause harm without asking leading questions with explicit examples of tasks that exhibit stereotypes or skews against non-binary genders. We distributed the survey through channels
like social media and mailing lists at universities and organizations. We had 19 individuals respond to our survey. While existing research has surveyed non-binary individuals on the harms of gendered web forms \citep{scheuerman2021}, there is no precedent for our survey on language technology harms, so our primary intent with this sample of respondents was to assess the efficacy of our survey design. 


\paragraph{Survey Structure}
The survey was anonymous, with no financial compensation, and questions were kept \emph{optional}. Further ethical considerations are presented in Section \ref{sec:ethics}.
In the following subsections, we briefly summarize our survey design and survey responses. We provide the full survey, 
our rationale for each question, and qualitative analysis of all responses received in Appendix \ref{sec:App survey}.


\subsubsection{Demographic information}

We asked survey respondents for demographic information to better understand the intersections of their identities. Demographic information included gender identity, ethnicity, AI experience, etc. 
 84.2\% of respondents use pronouns $they/them$, 26.3\% use $she/her$, 15.8\% use $he/him$, and 5.3\% use $xe/xem$. 31.6\% use multiple sets of pronouns. Additionally, an overwhelming majority (all but two) of our respondents identified as white and/or Caucasian. 
No respondents were Black, Indigenous, and/or Latinx, and two respondents were people of color. Furthermore, 52.6\% of respondents are originally from the US, 63.2\% current live in the US, 
and the majority of others are originally from or currently live in Canada and countries in Western Europe.  
This limits the conclusions we can reach from this sample's responses.
All respondents were familiar with AI, through their occupation, coursework, books, and social media 
(more details in Appendix \ref{app: deminfo}).

\subsubsection{Harms in Language Tasks}

This segment first defined representational and allocational harms~\cite{barocas2017problem} and introduced three common NLP tasks (Named Entity Recognition (NER), Coreference Resolution, and Machine Translation) using publicly-available AllenNLP demos \citep{Gardner2017AllenNLP}, which survey respondents engaged with to experiment with potential harms. The demos were accompanied by \emph{non-leading} questions about representational and allocational harms, if any, that non-binary communities could face as a result of these tasks. The questions were intentionally phrased to ask about the harms that could occur rather than imply likely harms (see Appendix \ref{sec:App survey}). We summarize the responses to these questions in Table \ref{tab:nlptasksharms}, where we see that misgendering of persons is a common concern across all three tasks. We found that, for all tasks, above 84\% of respondents could see/think of undesirable outcomes for non-binary genders. Furthermore, the severity of harms, as perceived by subjects of the survey, is the highest in machine translation, which is also a task more commonly used by the population at large.
We provide descriptions of the tasks and in-depth analyses of all the responses in Appendix \ref{sec:App nlpharms}.


\begin{table*}[t]
    \centering
    \small
  
    \begin{adjustbox}{width=\textwidth,totalheight=\textheight,keepaspectratio}
    \begin{tabular}{|p{0.08 \linewidth}|p{0.3 \linewidth}|p{0.31 \linewidth}|p{0.31 \linewidth}|}
    \hline
         & \textbf{Named Entity Recognition (NER)} & \textbf{Coreference Resolution} & \textbf{Machine Translation} \\
        \hline
        \centering \textbf{Example representational harms} & 
         \vspace{-4mm}
        \begin{itemize}[leftmargin=*]
            \item systematically mistags neopronouns and singular $they$ as non-person entities
            \item unable to tag non-binary chosen names as $Person$, e.g. the name ``A Boyd'' is not recognized as referring to a $Person$
            \item tags non-binary persons as $Person - male$ or $Person - female$
        \end{itemize} &
         \vspace{-4mm}
        \begin{itemize}[leftmargin=*]
            \item may incorrectly links $s/he$ pronouns with non-binary persons who do not use binary pronouns
            \item does not recognize neopronouns
            \item cannot link singular $they$ with individual persons, e.g. In ``Alice Smith plays for the soccer team. They scored the most goals of any player last season.'', $they$ is linked with $team$ instead of with $Alice$
        \end{itemize} & 
         \vspace{-3mm}
        \begin{itemize}[leftmargin=*]
            \item translates from a language where pronouns are unmarked for gender and picks a gender grounded in stereotypes associated with the rest of the sentence, e.g. translates ``(3SG) is a nurse'' (in some language) to ``She is a nurse'' in English 
             \vspace{-2mm}
            \item translates accepted non-binary terms in one language to offensive terms in another language, e.g. $kathoey$, which is an accepted way to refer to trans persons in Thailand, translates to $ladyboy$ in English, which is derogatory
        \end{itemize}\\
        \hline
        \centering \textbf{Example allocational harms} & 
         \vspace{-4mm}
        \begin{itemize}[leftmargin=*]
            \item NER-based resume scanning systems throw out resumes from non-binary persons for not having a recognizable name
            \item non-binary persons are unable to access medical and government services if NER is used as a gatekeeping mechanism on websites
            \item non-binary people with diverse and creative names are erased if NER is employed to build a database of famous people
        \end{itemize} & 
         \vspace{-4mm}
        \begin{itemize}[leftmargin=*]
            \item a coref-based ranking system undercounts a non-binary person's citations (including pronouns) in a body of text if the person uses $xe/xem$ pronouns
            \vspace{-2mm}
            \item a coref-based automated lease signing system populates referents with $s/he$ pronouns for an individual who uses $they/them$ pronouns, forcing self-misgendering
             \vspace{-2mm}
            \item a coref-based law corpora miner undercounts instances of discrimination against non-binary persons, which delays more stringent anti-discrimination policies
        \end{itemize} & 
        \vspace{-4mm}
        \begin{itemize}[leftmargin=*]
            \item machine-translated medical and legal documents applies incorrectly-gendered terms, leading to incorrect care and invalidation, e.g. a non-binary AFAB person is not asked about their pregancy status when being prescribed new medication if a translation system applies masculine terms to them
            \vspace{-1mm}
            \item machine-translated evidence causes non-binary persons to be denied a visa or incorrectly convicted of a crime
        \end{itemize}\\
        
        \hline
    \end{tabular}
    \end{adjustbox}
    \normalsize
    \vspace{-2mm}
    \caption{Summary of survey responses regarding harms in NLP tasks.}
    \label{tab:nlptasksharms}
    \vspace{-5mm}
\end{table*}

\subsubsection{Broader Concerns with Language Technologies}

This segment was purposely kept less specific to understand the harms in different domains (healthcare, social media, etc.) 
as perceived by different non-binary individuals.
We first list some domains to which language models can be applied along with summarized explanations of respondents regarding undesirable outcomes 
(see Appendix \ref{sec:App broader concerns} for in-depth analyses).

\noindent \emph{$\bullet$ Social Media:}
LGBTQ+ social media content is automatically flagged at higher rates. Ironically, language models can fail to identify hateful language targeted at non-binary people. Further, if social media sites attempt to infer gender from name or other characteristics, this can lead to incorrect pronouns for non-binary individuals. Additionally, ``language models applied in a way that links entities across contexts are likely to out and/or deadname people, which could potentially harm trans and non-binary people''. Moreover, social media identity verification could incorrectly interpret non-binary identities as fake or non-human.

\noindent \emph{$\bullet$ Healthcare:} Respondents said that ``healthcare requires engaging with gender history as well as identity'', which current language models are not capable of doing.
Additionally, language models could ``deny insurance claims, e.g. based on a `mismatch' between diagnosis and gender/pronouns''.

\noindent \emph{$\bullet$ Education:} Language models in automated educational/grading tools could ``automatically mark things wrong/`ungrammatical' for use of non-standard language, singular $they$, neopronouns, and other new un- or creatively gendered words''.

Additionally, respondents discussed some language applications that could exacerbate misgendering, non-binary erasure, transphobia, and the denial of cisgender privilege. 
Some examples were how automated summarization could fail to recognize non-binary individuals as people, language generation cannot generate text with non-binary people or language, speech-to-text services cannot handle neopronouns, machine translation cannot adapt to rapidly-evolving non-binary language, and
automated gender recognition systems only work for cis people (Appendix \ref{sec:App broader concerns}). 

The barriers \citep{barocas-hardt-narayanan} to better including non-binary persons in language models, as explained in the responses, are as follows 
(definitions and in-depth analyses in Appendix \ref{sec:App broader concerns}).

\noindent \emph{$\bullet$ Tainted Examples:} Since the majority of training data are scraped from sources like the Internet, which represent ``hegemonic viewpoints'', they contain few mentions of non-binary people; further, the text is often negative, 
and positive gender non-conforming content is not often published.

\noindent \emph{$\bullet$ Limited Features:} Data annotators may not recognize or pay attention to non-binary identities and may lack situational context.

\noindent \emph{$\bullet$ Sample Size Disparities:} Non-binary data may be ``discarded as `outliers''' and ``not sampled in training data'', non-binary identities may not be possible labels, developer/research teams tend to ``want to simplify variables and systems'' and may not consider non-binary persons prevalent enough to change their systems for.

\vspace{-1.5mm}
\subsection{Limitations and Future Directions}
We found that our survey, without any leading questions, was effective at getting respondents to recount language technology harms they had experienced on account of their gender, and brainstorm harms that could affect non-binary communities. However, our survey reaches specific demographics of ethnicity, educational background, etc. The responses 
equip us to better reach out to diverse groups of persons, including those without familiarity with AI and/or not fluent in English. Some respondents also indicated that language models could be used violently or to enable existing discriminatory policies, which should be explored in future related work.
Ultimately, we hope our survey design serves as a model for researching the harms technologies pose to marginalized communities.

\section{Data and Technical Challenges}
As a consequence of historical discrimination and erasure in society, narratives of non-binary persons are either largely missing from recorded text or have negative connotations. Language technologies also reflect and exacerbate these biases and harms, as discussed in Section \ref{sec:survey}, due to tainted examples, limited features, and sample size disparities. These challenges are not well understood. We discuss the different fundamental problems that need to be acknowledged and addressed to strategize and mitigate the cyclical erasure and misgendering of persons as a first step towards building language models that are more inclusive.

\subsection{Dataset Skews}

\label{sec: data skew}
The large text dumps often used to build language representations have severe skews with respect to gender and gender-related concepts. Just observing pronoun usage, English Wikipedia text (March 2021 dump), which comprises $4.5$ billion 
tokens, has over $15$ million mentions of the word $he$, $4.8$ million of $she$, $4.9$ million of $they$,  $4.5$ thousand of $xe$,  $7.4$ thousand of $ze$, and $2.9$ thousand of $ey$. Furthermore, the usages of non-binary pronouns\footnote{Neopronouns and gendered pronouns not ``he'' or ``she''} were mostly not meaningful with respect to gender (Appendix \ref{App: Dataset skews}). $Xe$, as we found by annotation and its representation, is primarily used as the organization $Xe$ rather than the pronoun $xe$. $Ze$ was primarily used as the Polish word $that$, as indicated by its proximity to mostly Polish words like $nie$, i.e. $no$, in the GloVe representations of the words, and was also used for characterizing syllables. Additionally, even though the word $they$ occurs comparably in number to the word $she$, a large fraction of the occurrences of $they$ is as the plural pronoun, rather than the singular, non-binary pronoun $they$. 
Some corpora do exist such as the Non-Binary Wiki\footnote{\url{https://nonbinary.wiki/wiki/Main\_Page}} which contain instances of meaningfully used non-binary pronouns. However, with manual evaluation, we see that they have two drawbacks: (i) the narratives are mostly short biographies and lack the diversity of sentence structures as seen in the rest of Wikipedia, and (ii) they have the propensity to be dominated by Western cultures, resulting in further sparsification of diverse narratives of non-binary persons. 


\subsection{Text Representation Skews}
\label{sec: representation}
Text representations have been known to learn and exacerbate skewed associations and social biases from underlying data~\cite{ZhaoWYOC17, bender2021dangers,dev2020geometry}, thus propagating representational harm. We examine representational skews with respect to pronouns and non-binary-associated words that are extremely sparsely present in text. 









\noindent \paragraph{\textbf{Representational erasure in GloVe.}}

\begin{table}[t]
\centering
\small
\begin{tabular}{|l|p{5cm}|}\hline
\textbf{Pronoun} & \textbf{Top 5 Neighbors}\\\hline
He& \textit{`his', `man', `himself', `went', `him'}\\
She& \textit{`her', `woman', `herself', `hers', `life'}\\
They& \textit{`their', `them', `but', `while', `being'}\\
Xe& \textit{`xa', `gtx', `xf', `tl', `py'}\\
Ze & \textit{`ya', `gan', `zo', `lvovic', `kan'}\\\hline
\end{tabular}
\normalsize
\vspace{-2mm}
\caption{Nearest neighbor words in GloVe for binary and non-binary pronouns.}
\label{tbl: nearest nbrs}
\vspace{-4.5mm}
\end{table}
Table \ref{tbl: nearest nbrs} shows the nearest neighbors of different pronouns in their GloVe representations trained on English Wikipedia data. The singular pronouns $he$ and $she$ have semantically meaningful neighbors as do their possessive forms (Appendix \ref{app: representational skews}).  The same is not true for non-binary neopronouns $xe$ and $ze$ which are closest to acronyms and Polish words, respectively. These reflect the disparities in occurrences we see in Section \ref{sec: data skew} and show a lack of meaningful encodings of non-binary-associated words. 


\noindent \paragraph{\textbf{Biased associations in GloVe.}}

Gender bias literature primarily focuses on stereotypically gendered occupations~\cite{bolukbasi2016man,De-artega2019bias}, with some exploration of associations of binary gender and adjectives  ~\cite{dev2019attenuating, Caliskan183}. While these associations are problematic, there are additional, significantly different biases against non-binary genders,
namely misrepresentation and under-representation. Furthermore, non-binary genders suffer from a sentiment (positive versus negative) bias.
Gender-occupation associations are not a dominant stereotype observed across all genders (Table \ref{tbl: occupations}), where non-binary words like $transman$ and $nonbinary$ are not dominantly associated with either stereotypically male or female occupations. In fact, most occupations exhibit no strong correlation with words and pronouns associated with non-binary genders (see Appendix \ref{app: representational skews}). 



\begin{table}[]
\small
		\begin{tabular}{|c|c|c|c|c|}
		\hline
		\textbf{Word} & \textbf{Doctor} & \textbf{Engineer} & \textbf{Nurse} & \textbf{Stylist}  \\
			\hline
			man & 0.809 & 0.551 & 0.616 & 0.382   \\
			woman & 0.791 & 0.409 & 0.746 & 0.455  \\
			transman & -0.062 & -0.152  & -0.095 & 0.018 \\
			transwoman & -0.088 & -0.271 & 0.050 &  0.062 \\
			 nonbinary & 0.037 & -0.243 & 0.129 & 0.015\\
			\hline
		\end{tabular}
	\normalsize
	\vspace{-2mm}
	\caption{Cosine similarity: gendered words vs common occupations.}
	\label{tbl: occupations}
	\vspace{-6mm}
\end{table}

To investigate sentiment associations with binary versus non-binary associated words, we use the WEAT test~\cite{Caliskan183} with respect to pleasant and unpleasant attributes (listed in Appendix  \ref{app: WEAT}).
Since neopronouns are not well-embedded, we compare disparate sentiment associations between binary versus non-binary pronouns, gendered words and proxies (e.g., $male$, $female$ versus $transman$, $genderqueer$, etc.). The WEAT score is 0.916, which is non-zero, i.e. ideal, significantly large (detailed analysis in Appendix \ref{app: WEAT}), and indicates disparate sentiment associations between the two groups. 
 For $man$ and $woman$, the top nearest neighbors include $good$, $great$ and $good$, $loving$, respectively. However, for $transman$ and $transwoman$, top words include $dishonest$, $careless$ and $unkind$, $arrogant$. This further substantiates the presence of biased negative associations, as seen in the WEAT test.
Furthermore, the nearest neighbors of words associated with non-binary genders 
 are derogatory (see Appendix Table \ref{App: derogatory}). In particular, $agender$ and $genderfluid$ have the neighbor $negrito$, meaning ``little Black'', 
while $genderfluid$ has $Fasiq$, which is an Arabic word used for someone 
of corrupt moral character.





\noindent \paragraph{\textbf{Representational erasure in BERT.}}
\footnote{Code and supporting datasets can be found at https://github.com/uclanlp/harms-challenges-non-binary-representation-NLP}
Pronouns like $he$ or $she$ 
are  part of the word-piece embedding vocabulary that composes the input layer in BERT. However, similar length neo-pronouns $xe$ or $ze$ are deemed as out of vocabulary by BERT, indicating infrequent occurrences of each word and a relatively poor embedding. 

BERT's contextual representations should ideally be able to discern between singular mentions of $they$ (denoted $they(s)$) and plural mentions of $they$ (denoted $they(p)$), and to some extent it indeed is able to do so, but not with high accuracy.
%
For this, we train BERT as a classifier to disambiguate between singular and plural pronouns. Given a sentence containing a masked pronoun along with two proceeding sentences, it predicts whether the pronoun is singular or plural. We build two separate classifiers $C_1$ and $C_2$. Both are first trained on a dataset containing sentences with $i$ or $we$ (singular versus plural; details on this experiment in Appendix \ref{app: bert}).
Next, $C_1$ is trained on classifying  $they(s)$ vs $they(p)$ 
while $C_2$ is trained on classifying  $he$ vs $they(p)$. This requires balanced, labeled datasets for both classifiers. The text spans for $they(p)$ are chosen randomly from Wikipedia containing pairs of sentences such that the word $they$ appears in the second sentence (with no other pronoun present) and the previous sentence has a mention of two or more persons (determined by NER). This ensures that the word $they$ in this case was used in a plural sense. 
Since Wikipedia does not have a large number of sentences using $they(s)$,
for such samples, we randomly sample them from Non-Binary Wiki (Section \ref{sec: data skew}). 
The sentences are manually annotated for further confirmation of correct usage of each pronoun.
We follow the procedure of data collection for $they(s)$ to create datasets for sentences using the pronoun $he$ from Wikipedia. Therefore, while $C_1$ sees a dataset containing samples with $they(s)$ or $they(p)$, $C_2$ sees samples with $he$ or $they(p)$. In each dataset, however, we replace the pronouns with the $[MASK]$ token.
We test $C_1$ and $C_2$ on their ability to correctly classify a new dataset for $they(p)$ (collected the same way as above). If $C_1$ and $C_2$ learn the difference between the singular and plural representations, each should be able to classify all sentences as plural with net accuracy 1. While the accuracy of $C_2$ is $83.3\%$, $C_1$'s accuracy is only $67.7\%$.
This indicates $they(s)$ is not as distinguishable from $they(p)$ as a binary-gendered pronoun (further experiments are in Appendix \ref{app: bert}).
\noindent \paragraph{\textbf{Biased representations with BERT.}}
 To understand biased associations in BERT, we must look at representations of words with context. For demonstrating skewed associations with occupations (as shown for GloVe), we adopt the sentence template ``[pronoun] is/are a [target].''. 
We iterate over a commonly-used list of popular occupations~\cite{dev2019measuring}, broken down into stereotypically female and male~\cite{bolukbasi2016man}.  
We get the average probability for predicting each gendered pronoun (Table \ref{tab:bert1}) $P([pronoun]|[target]=occupation)$ over each group of occupations. 
The results in Table \ref{tab:bert1} demonstrate that the occupation biases in language models with respect to binary genders is not meaningfully applicable for all genders.  

\begin{table}[]
    \centering
    \small
    \begin{tabular}{|c|ccc|}
    \hline
        \textbf{Pronouns} & \multicolumn{3}{c|}{\textbf{Occupations Categories}} \\
       & Male & Female & All \\
        \hline
         he & 0.5781 & 0.1788 & 0.5475  \\
         she & 0.1563 & 0.4167 & 0.2131 \\
         they & 0.1267 & 0.1058 & 0.1086\\
         xe &  2.1335e-05 & 1.9086e-05 & 1.6142e-5 \\
         ze & 7.4232e-06 & 6.0601e-06  & 5.6769e-6 \\ \hline
    \end{tabular}
    \normalsize
    \vspace{-1mm}
    \caption{Pronoun associations with (i) stereotypically male, (ii) stereotypically female, and (iii) extensive list of 180 popular occupations. Values are aggreagated probabilities (higher value implies more associated; see main text for more details).}
    \label{tab:bert1}
    \vspace{-5mm}
\end{table}

\noindent \paragraph{\textbf{BERT and Misgendering.}}

Misgendering is a harm experienced by non-binary persons, as emphasized in the survey (see Section \ref{sec:survey}). Further, misgendering in language technologies can reinforce erasure and the diminishing of narratives of non-binary persons. We propose an evaluation framework here that demonstrates how BERT propagates this harm. 
We set up sentence templates as such:

[\textbf{Alex}] [\textbf{went to}] the [\textbf{hospital}] for \underline{[PP]} [\textbf{appointment}]. \underline{[MASK]} was [\textbf{feeling sick}].

Every word within [] is varied. The words in bold are varied to get a standard set of templates (Appendix \ref{app: bert}). These include the verb, the subject, object and purpose. We iterate over $919$ names available from SSN data which were unisex or least statistically associated with either males or females~\cite{538unisex}. We choose this list to minimize binary gender correlations with names in our test.
Next, we vary the underlined words in pairs. The first of each pair is a possessive pronoun (PP) which we provide explicitly (thus indicating correct future pronoun usage) and use BERT to predict the masked pronoun in the second sentence in each template. The ability to do so for the following five pairs is compared: (i) $his$, $he$ (ii) $her$, $she$ (iii) $their$, $they$ (iv) $xir$, $xe$ and (v) $zir$, $ze$ in Table \ref{tab:bert3}, where \emph{Accuracy} is the fraction of times the correct pronoun was predicted with highest probability and the score \emph{Probability} is the average probability associated with the correct predictions.  The scores are high for predicting $he$ and $she$, but drop for $they$. For $xe$ and $ze$ the amount by which the accuracy drops is even larger, but we can attribute this to the fact that these neopronouns are considered out of vocabulary by BERT. This demonstrates how models like BERT can explicitly misgender non-binary persons even when context is provided for correct pronoun usage.




\begin{table}[]
    \centering
    \small
    \begin{tabular}{|c|c|c|}
    \hline
        \textbf{Pronouns pairs} & \textbf{Accuracy} & \textbf{Probability} \\
        \hline
         his-he & 0.861 & 0.670  \\
         her-she & 0.785 & 0.600 \\
         their-they & 0.521 & 0.391 \\
         xir-xe & 0.0 & 1.137e-05 \\
         zir-ze & 0.0 & 1.900e-04 \\
         \hline
    \end{tabular}
    \normalsize
    \vspace{-2mm}
    \caption{BERT performance for gendered pronoun predictions. Accuracy is the fraction of times the correct pronoun was predicted and probability is the aggregated probability associated with correct prediction.}
    \label{tab:bert3}
    \vspace{-5mm}
\end{table}

\section{Discussion and Conclusion}


This work documents and demonstrates specific challenges towards making current language modeling techniques inclusive of all genders and reducing the extent of discrimination, misgendering, and cyclical erasure it can perpetrate.  
In particular, our survey identifies numerous representational and allocational harms, as voiced from individuals affected, and we demonstrate and quantify several cases where the roots of these concerns arise in popular language models.  
Some efforts in the NLP community have worked towards countering problems in task-specific data sets with skewed gender tags, due to underrepresentation of non-binary genders. Notably, \citet{cao2020towards} and \citet{10.1162/coli_a_00413} introduce a gender-inclusive dataset GICoref for coreference resolution and \citet{sun2021they} propose rewriting text containing gendered pronouns with $they$ as the substituted pronoun to obtain more gender-neutral text. 
The challenges still remain that (i) not all neopronouns will have sufficient data in real-world text, and (ii) considering non-binary genders as a monolithic third category 
(i.e. male, female, and gender-neutral) is counter-productive and perceived as harmful (Section \ref{sec:survey}).  
While these efforts are a start in moving away from binary gender, it is questionable if gender should be defined as discrete quantities in language modeling, when in reality, it is of a fluid nature.  Furthermore, models currently do not account for the mutability of gender and the language around it, and even if they did, they would likely assume there exist well-defined points at which individuals and words transition, which too is detrimental (as documented in our survey, see Section \ref{sec:survey}). 
Representing gender is as complex as the concept of gender itself. Bucketing gender in immutable, discrete units and trying to represent each, would inevitably result in marginalization of sections of the population to varied extents. 
As our survey catalogs how pronounced the harms of being consistently misgendered and diminished are, we encourage future work to carefully examine how (\emph{and if}) to define and model gender in language representations and tasks.

This work sets the interdisciplinary stage for rethinking and addressing challenges with inclusively modeling gender in language representations and tasks. Any viable solution cannot simply be a quick fix or patch, but must rely on a bottom-up approach involving affected persons system-wide, such as in annotation and human-in-the-loop mechanisms. Simultaneously, research into monitoring language technologies over time to detect harms against non-binary individuals is critical. It is further paramount to transparently communicate the performance of language technologies for non-binary persons and possible harms. In the case of harm, non-binary individuals must be able to obtain valid recourse to receive a more favorable outcome, as well as have the opportunity to provide feedback on the model's output and have a human intervene.

\paragraph{\large{Acknowledgements}}
\normalsize
We would like to thank Emily Denton and Vasundhara Gautam for their extensive feedback on our survey and our anonymous reviewers for detailed feedback on the paper. Furthermore, we would like to thank and deeply appreciate all the survey respondents for the time and effort they invested into drawing from their lived experiences to provide innumerable informative insights, without which this project would not have been possible. We would also like to thank Vasundhara Gautam, Gauri Gupta, Krithika Ramesh, Sonia Katyal, and Michael Schmitz for their feedback on drafts of this paper.


This work was supported by NSF grant \#2030859 to the Computing Research Association for the CIFellows Project. Additionally, we also thank the support from Alfred P. Sloan foundation, NSF IIS-1927554, NSF CCF-1350888, CNS-1514520, CNS-1564287, IIS-1816149, and Visa Research.

\section{Broader Impact and Ethics}
\label{sec:ethics}
Our survey was reviewed by an Institutional Review Board (IRB), and the IRB granted the survey Exempt status, requiring only signed informed consent for the entire study, as the survey posed minimal risk, was conducted online, and did not involve treating human subjects. In the survey, we ask non-leading questions to record perceptions of participants and not preemptively impose possibilities of harm. Questions were also optional to enable participants to control the amount of emotional labor they incurred while taking the survey. We were cautious to protect identities of persons who took the survey; we analyzed aggregated data and any quotes of text analyzed or mentioned cannot be traced back to an individual. 

Inclusivity and fairness are important in NLP and its wide ranging applications. Gender, when modeled in these applications has to reflect fairly the concepts of gender identity and expression. Failure to do so leads to severe harms, especially for persons not subscribing to binary gender. In this work, we attempt to broaden the awareness of gender disparities and motivate future work to discuss and further address the harms propagated by language technologies. We emphasize the importance of centering the lived experiences of marginalized communities therein.

\bibliography{custom}
\bibliographystyle{acl_natbib}

\clearpage
\newpage

\pagenumbering{arabic}
\setcounter{table}{0}

\twocolumn[
\begin{center}
{\Large \textbf{\\ Appendix: Harms of Gender Exclusivity and Challenges in Non-Binary Representation in Language Technologies \\ \vspace{1in}}}
\end{center}
]

\appendix

\section{Survey}
\label{sec:App survey}

Below, we provide the full Survey on Harms, our rationale for each question, and qualitative analysis of all responses received.

\subsection{Demographic information}
\label{app: deminfo}

\noindent \textbf{Q: } What pronouns do you use? (checkboxes) \\
\textbf{Options:} they/them, she/her, he/him, xe/xem, e/em, ze/hir, I don’t use pronouns, I am questioning about my pronouns, I don’t want to answer this question, Other (option to specify in text field) \\

\begin{table}[h]
    \centering
    \begin{tabular}{|c|c|}
    \hline
        \textbf{Pronouns} & \textbf{Percentage of Total Respondents} \\
        \hline
         they/them & 84.2\%  \\
         she/her & 26.3\% \\
         he/him & 15.8\% \\
         xe/xem & 5.3\% \\
         \hline
    \end{tabular}
    \caption{Survey Pronouns Distribution}
    \label{tab:pronouns}
\end{table}

\noindent For this question, we ensured to allow respondents to check multiple options, as many non-binary persons use more than one set of pronouns. Table \ref{tab:pronouns} summarizes the distribution of pronouns. 31.6\% of respondents used more than one set of pronouns (e.g. $she/her$ and $xe/xem$). \\

\noindent \textbf{Q: } What are your pronouns in other languages? (text field)  \\

\noindent We collected this information because in languages without referential gender or where pronouns are seldom used, pronouns can be less central to one's gender identity; thus, we wanted to discover pronouns non-binary persons use in languages other than English. These data could be useful to future research on the harms of non-English language technologies to non-binary communities.

Many respondents listed their pronouns in other languages. Unmodified responses included:
\begin{quote}
    hen (swedish), iel/any pronoun (french)
\end{quote}
\begin{quote}
    Ta (Mandarin), sie (German)
\end{quote}
\begin{quote}
    nin/nim (German)
\end{quote}
\begin{quote}
   ``Hen'' in Swedish, ``hän'' in Finnish, none in Japanese (pronouns are seldom used at all). Unfortunately, many languages still do not have more commonly accepted gender-neutral pronouns, or I'd use them.
\end{quote}

\noindent \textbf{Q: } What is your sexual orientation? (checkboxes) \\
\textbf{Options:} Lesbian, Gay, Bisexual, Asexual, Pansexual, Queer, Straight, Questioning, I don’t want to answer this question, Other (option to specify in text field) \\

\begin{table}[]
    \centering
    \begin{tabular}{|c|c|}
    \hline
    \textbf{Sexual} & \textbf{Percentage of} \\
    \textbf{Orientation} & \textbf{Total Respondents} \\
        \hline
         Queer & 57.9\%  \\
         Bisexual & 26.3\% \\
         Asexual & 26.3\% \\
         Pansexual & 26.3\% \\
         Straight & 10.3\% \\
         Gay & 5.3\% \\
         Questioning & 5.3\% \\
         I don’t want to answer this question & 5.3\%  \\
         \hline
    \end{tabular}
    \caption{Survey Sexual Orientation Distribution}
    \label{tab:sexuality}
\end{table}

\noindent For this question, we ensured to allow respondents to check multiple options. Results may be found in Table  \ref{tab:sexuality}. \\

\noindent \textbf{Q: } What is your gender? (checkboxes) \\
\textbf{Options:} Man, Woman, Non-binary, Genderqueer, Third-gender, Genderfluid, Gender non-conforming, Pangender, Two-Spirit, Agender, Questioning, I don’t want to answer this question, Other (option to specify in text field) \\

\begin{table}[]
    \centering
    \begin{tabular}{|c|c|}
        \hline
        \textbf{Gender} & \textbf{Percentage of} \\
        & \textbf{Total Respondents} \\
        \hline
         Non-binary & 73.7\% \\
         Genderqueer & 31.6\% \\
         Agender & 31.6 \% \\
         Gender non-conforming & 21.1\% \\
         Questioning & 21.1\% \\
         Woman & 15.8\% \\
         Man & 10.5\% \\
         Genderfluid & 5.3\% \\
         demi-boy  & 5.3\% \\
         \hline
    \end{tabular}
    \caption{Survey Gender Distribution}
    \label{tab:gender}
\end{table}

\noindent For this question, we ensured to allow respondents to check multiple options. Results may be found in Table \ref{tab:gender}. Table \ref{tab:gender} demonstrates that many individuals identify with multiple genders and that we achieved a Western gender-diverse sample. \\

\noindent \textbf{Q: } In a few words, how would you describe your gender and sexual orientation? If you feel that the above questions are not able to capture your gender and sexual orientation, feel free to use this question to enter your response. (text field) \\

\noindent We achieved a Western gender and sexuality-diverse sample. Unmodified responses included:
\begin{quote}
   agender gray-asexual, falling under the broader categories of nonbinary, trans, asexual, and queer
\end{quote}

\begin{quote}
   I am a bisexual demi-boy existing somewhere between male and a third gender space
\end{quote} 

\begin{quote}
   gender non-conforming ace; sex-averse; at my happiest with intimate- or queer-platonic relationships which are generally not labeled ``romantic''
\end{quote} 
\begin{quote}
   I identify as pansexual and queer. I'm questioning my gender but I'm likely somewhere between nonbinary woman and agender.
\end{quote} 
\begin{quote}
   I present as somewhat femme to “tomboy” and generally live life as a woman, but internally am very agender. I wouldn’t care about my pronouns, but I feel obligated to state them to support various groups of people and generally support the visibility of women. And I’m a pan-romantic sex-disinterested grey-asexual.
\end{quote} 
\begin{quote}
   I'm panromantic asexual and I'm coming to the conclusion that I'm also agender.
\end{quote}

\noindent \textbf{Q: } Are you trans? (radio buttons) \\
\noindent \textbf{Options: } Yes, No, I am questioning about my gender, I don’t want to answer this question \\

\begin{table}[]
    \centering
    \begin{tabular}{|c|c|}
        \hline
        \textbf{Trans Identification} & \textbf{Percentage of} \\
        & \textbf{Total Respondents} \\
        \hline
         Yes & 63.2\%  \\
         I am questioning about my gender & 21.1\% \\
         No & 10.5\% \\
         I don’t want to answer this question & 5.3\% \\
         \hline
    \end{tabular}
    \caption{Survey Trans Identification Distribution}
    \label{tab:trans}
\end{table}

\noindent Results may be found in Table \ref{tab:trans}. \\

\noindent \textbf{Q: } In a few words, how would you describe your ethnicity? (text field) \\

\noindent \textbf{Q: } Are you Black, Latinx, and/or Indigenous? (radio buttons) \\
\noindent \textbf{Options: } Yes, No, I don’t want to answer this question \\

\noindent \textbf{Q: } Are you a person of color? (radio buttons) \\
\noindent \textbf{Options: } Yes, No, I don’t want to answer this question \\

\noindent \textbf{Q: } Which country are you from originally? (text field) \\

\noindent \textbf{Q: } Which country do you live in? (text field) \\

\noindent We intentionally made many of the above questions free-response to allow respondents to explain their ethnicity and nationality.

An overwhelming majority (all but two) of our respondents identified as white, white British, Western European, and/or Caucasian. No respondents were Black, Indigenous, and/or Latinx, and two respondents were people of color.

Furthermore, 52.6\% of respondents are originally from the US, and 63.2\% current live in the US. Two respondents grew up in India and Singapore, and Taiwan. The remaining respondents are originally from or currently live in Canada and countries in Western Europe, like France, the UK, Germany, and Sweden.

This severely limits the conclusions we can reach from this sample's responses. In the future, we will improve survey outreach to diversify our sample. \\

\noindent \textbf{Q: } How would you describe your occupation? (text field) \\

\noindent \textbf{Q: } How would you describe your familiarity with AI? (text field) \\

\begin{table}[]
    \centering
    \begin{tabular}{|c|c|}
        \hline
        \textbf{Occupation} & \textbf{Percentage of} \\
        & \textbf{Total Respondents} \\
        \hline
         Researcher & 42.1\% \\
         Software Engineer & 31.6\% \\
         Student & 26.3\% \\
         Unemployed & 5.3\% \\
         \hline
    \end{tabular}
    \caption{Survey Occupation Distribution}
    \label{tab:occupation}
\end{table}

\noindent Occupation results may be found in Table \ref{tab:occupation}. All respondents were familiar with AI, through their occupation, coursework, books they had read, and/or social media. We recognize that these occupations and level of familiarity of AI are correlated with privilege and socioeconomic status; in the future, we will expand our sample beyond those who work in/are familiar with AI.

\subsection{Harms in Language Tasks}
\label{sec:App nlpharms}
    
This segment first defined representational and allocational harms, and then introduced three tasks (Named Entity Recognition, Coreference Resolution, and Machine Translation) using publicly-available demos \citep{Gardner2017AllenNLP} which survey subjects engaged with. The demos were accompanied with non-leading questions about potential representational and allocational harms that non-binary communities could face as a result of these tasks. \\

\noindent \textbf{Named Entity Recognition (NER)} involves taking an unannotated block of text, such as this one: ``Microsoft acquired GitHub in 2018'', and producing an annotated block of text that highlights the names of entities: ``[Microsoft]$_{\text{Organization}}$ acquired [GitHub]$_{\text{Organization}}$ in [2018]$_{\text{Time}}$''. We provided survey participants with the AllenNLP Named Entity Recognition \footnote{\url{https://demo.allennlp.org/named-entity-recognition/fine-grained-ner}} demo \citep{Gardner2017AllenNLP}. \\

\noindent \textbf{Q: } Can you see/think of cases where Named-Entity Recognition with current language models could have an undesirable outcome for non-binary genders? (radio buttons) \\
\noindent \textbf{Options: } Yes, No \\

\noindent 84.2\% of respondents indicated Yes, while 15.8\% indicated No. \\

\noindent \textbf{Q: } Does it cause representational harm? Please provide an example(s). (text field) \\

\noindent \textbf{Q: } What’s the severity of the representational harm? (1-5 scale) \\

\noindent Respondents argued that NER ``systematically mistags neopronouns, which reinforces the stereotype that neopronouns/alternate pronouns are `hard' or `complicated' and is thus directly harmful to non-binary people''. Additionally, NER can assume singular ``they'' refers to multiple people, and it may label those who use ``it/its'' pronouns as objects, which is ``dehumanizing and reinforces a negative stereotype of non-binary persons''.

Another concern respondents raised was NER's inability to recognize the names of non-binary persons and correctly tag the Person entity, since many Western non-binary chosen names are creative and diverse, ``overlapping with common nouns'' (especially nature-related nouns), having ``uncommon orthographic forms'', and/or consisting of a single letter. For example, the AllenNLP NER demo cannot correctly tag the full name of a person in the case of a single-letter first name. NER performing badly on these names would ``reinforce that non-binary names are `weird'''.

Finally, respondents mentioned that NER systems that classify human entities as `Person-male' or `Person-female' and reinforce the gender binary can be psychologically harmful.

Overall, on a scale of 1-5, where 1 indicates ``No impact on lives'', 3 indicates ``Noticeably negatively affects lives'', and 5 indicates ``Significantly hinders lives'', 47.1\% of respondents said that the severity of NER's representational harm is a 3, 23.5\% said 2, 17.6\% said 4, 5.9\% said 1, and 5.9\% said 5. \\

\noindent \textbf{Q: } Can it cause allocational harm? Please provide an example(s) of use cases and the resultant allocational harm. (text field) \\

\noindent \textbf{Q: } What’s the severity of the allocational harm? (1-5 scale) \\

\noindent Respondents said that NER systems can be devastating when they are unable to recognize non-binary chosen names. For example, if organizations scan resumes using NER systems, job and fellowship applications from non-binary persons may be thrown out for ``not having a name''. Additionally, if NER systems are used for identity verification, non-binary persons could be ``systemically incorrectly labeled by these systems, which could come into play when a system that wants to verify the identity of an account concludes the account does not belong to a human''. Similarly, non-binary people may be unable to access medical and government-administered services if NER is used as a gatekeeping mechanism on healthcare and government websites. NER systems may also be used to automatically build a database of famous people from text data, and if non-binary names are less likely to be correctly recognized, they will be excluded from the database, which could exacerbate erasure.

Overall, on a scale of 1-5, where 1 indicates ``No impact on lives'', 3 indicates ``Noticeably negatively affects lives'', and 5 indicates ``Significantly hinders lives'', 25\% of respondents said that the severity of NER's allocational harm is a 3, 25\% said 2, 18.8\% said 4, 18.8\% said 5, and 12.5\% said 1. \\

\noindent \textbf{Coreference Resolution} is the task of finding all expressions that refer to the same entity in a block of text. For example, a coreference resolution system would determine that in ``[UCLA] is a public university. [It] offers courses in Computer Science.'', ``UCLA'' corefers with ``it''. It is an important step for a lot of higher level NLP tasks that involve natural language understanding such as document summarization, question answering, and information extraction. We provided survey participants with the AllenNLP Coreference Resolution \footnote{\url{https://demo.allennlp.org/coreference-resolution}} demo \citep{Gardner2017AllenNLP}. \\

\noindent \textbf{Q: } Can you see/think of cases where Coreference Resolution with current language models could have an undesirable outcome for non-binary genders? (radio buttons) \\
\noindent \textbf{Options: } Yes, No \\

\noindent 84.2\% of respondents indicated Yes, while 15.8\% indicated No. \\

\noindent \textbf{Q: } Does it cause representational harm? Please provide an example(s). (text field) \\

\noindent \textbf{Q: } What’s the severity of the representational harm? (1-5 scale) \\

\noindent Respondents argued that ``the potential of accidental misgendering is high''. For instance, coreference resolution systems could ``apply `s/he' to individuals who might not identify that way''. Furthermore, coreference resolution systems might ``incorrectly apply non-binary pronouns to people who do not use them, like applying `it' to a trans woman who uses `she' pronouns''; this would ``echo the societal harm in which people with nonstandard gender presentations are treated as less than human''.

Additionally, respondents mentioned that ``neopronoun users as a group are diminished when software does not work on language referencing them'', especially since neopronouns are often underrepresented or even non-existent in textual data. Erasing and neglecting neopronouns contribute to queer erasure, and ``when we build coreference systems that cannot handle neopronouns, we reinforce the stereotype that neopronouns/alternate pronouns are `hard' or `complicated', which is directly harmful to non-binary people''.

Similarly, a non-binary person referred to by name and then subsequently by ``they/them'' or ``it/its'' pronouns ``might fail to be identified as referring to the same person'', because coreference resolution systems could erroneously assume the person is multiple people or an object. For example, the AllenNLP coreference resolution demo cannot correctly handle singular ``they'' pronouns. One respondent found that in the example, ``Alice Smith plays for the soccer team. They scored the most goals of any player last season,'' the model connects ``they'' with ``team''; however, English speakers would be able to disambiguate and understand that ``they'' actually refers to ``Alice''.

Furthermore, respondents emphasized that coreference resolution systems can reinforce the idea that names/occupations/roles are gendered and that there are only two genders, e.g. `doctor' is much more likely to link to `he' than `they', `she', `xe', etc.

Overall, on a scale of 1-5, where 1 indicates ``No impact on lives'', 3 indicates ``Noticeably negatively affects lives'', and 5 indicates ``Significantly hinders lives'', 47.1\% of respondents said that the severity of coreference resolution's representational harm is a 4, 23.5\% said 2, 23.5\% said 3, 5.9\% said 5, and 0\% said 1. \\

\noindent \textbf{Q: } Can it cause allocational harm? Please provide an example(s) of use cases and the resultant allocational harm. (text field) \\

\noindent \textbf{Q: } What’s the severity of the allocational harm? (1-5 scale) \\

\noindent Respondents provided numerous realistic harmful use cases of coreference resolution. For instance, a ``ranking system where you count citations of a person from a body of text (including references to their pronouns which you would resolve through coreference resolution) could miss a lot of instances of people being cited with `xe/xem' pronouns, which would give them a lower ranking''. Another respondent conceived an example in which ``a person using singular `they' pronouns who was required to sign a lease that populated referents with `s/he' pronouns instead is forced to sign an incorrect acknowledgement or not obtain housing''. Furthermore, coreference resolution systems might cause applications for financial aid targeted solely at individuals from non-binary persons who use ``they/them'' pronouns  to be automatically flagged as ineligible. Finally, a respondent described a situation in which ``if coreference resolution is used to sort through large law corpora to find instances of non-binary people being discriminated against to see if more stringent policy should be put in place to stop discrimination, it may erroneously find that there are not many cases of this since `they' is not often linked to a specific person''.

Overall, on a scale of 1-5, where 1 indicates ``No impact on lives'', 3 indicates ``Noticeably negatively affects lives'', and 5 indicates ``Significantly hinders lives'', 35.7\% of respondents said that the severity of coreference resolution's allocational harm is a 4, 21.4\% said 2, 21.4\% said 3, 14.3\% said 5, and 7.1\% said 1. \\

\noindent \textbf{Machine Translation} systems translate text from one language to another. When translating from languages with pronouns that do not carry gender information (e.g. Tagalog) to those that have gendered pronouns (e.g. English), translation systems may impose incorrect binary pronouns on individuals. This can be problematic in several ways such as reinforcing gender stereotypes, and misgendering and excluding non-binary persons. We provided survey participants with Google Translate \footnote{\url{https://translate.google.com/}}. \\

\noindent \textbf{Q: } Can you see/think of cases where Machine Translation with current language models could have an undesirable outcome for non-binary genders? (radio buttons) \\
\noindent \textbf{Options: } Yes, No \\

\noindent 89.5\% of respondents indicated Yes, while 10.4\% indicated No. \\

\noindent \textbf{Q: } Does it cause representational harm? Please provide an example(s). (text field) \\

\noindent \textbf{Q: } What’s the severity of the representational harm? (1-5 scale) \\

\noindent Respondents overwhelmingly discussed the harm of machine translation systems ``translating from a language where pronouns are unmarked for gender, and picking a gender grounded in stereotypes associated with the rest of the sentence'' in the translation in the target language. Many respondents raised the example of translating ```3SG is a nurse' (in some language) to `She is a nurse' in English and `3SG is a doctor' (in some language) to `He is a doctor' in English''. Another example, based in heteronormativity, is Google Translate French-to-English ``translates `sa femme' (his/her/their wife) as `his wife' and `son mari' (his/her/their husband) as `her husband' even in sentences with context, e.g. `Elle et sa femme se sont mariées hier' (`she and her wife got married yesterday') is translated as `she and his wife got married yesterday'''.

Furthermore, the long-established gender-neutral pronouns ```hen' and `hän' from Swedish and Finnish'' and ``strategies to mix gendered inflections'' all often ``automatically translate to `her' or `him''' in English. In addition, machine translation systems can ``misinterpret non-binary names and  pronouns as referring to objects, thereby dehumanizing non-binary people''. This can lead to nonbinary people being misgendered if their pronouns do not align with the ones that the machine translation system imposed upon them. Moreover, ``if neopronouns are not even represented, then this also contributes to erasure of queer identity''; it is likely that neopronouns ``are represented as unknown tokens,'' which can be problematic. 

Additionally, many grammatically-gendered languages lack non-binary gender options, so a person may have their gender incorrectly ``binary-ified'' in the target language, which constitutes misgendering and ``is hurtful''.

Finally, differences in how languages talk about non-binary people are ``extremely nuanced'', which can lead to ``extremely disrespectful'' translations. One respondent explained that, while ``a common and accepted way to refer to trans people in Thailand is the word `kathoey', which translates to `ladyboy''', if someone called this respondent a ``ladyboy'' in English, the respondent would be extremely offended.

Overall, on a scale of 1-5, where 1 indicates ``No impact on lives'', 3 indicates ``Noticeably negatively affects lives'', and 5 indicates ``Significantly hinders lives'', 47.1\% of respondents said that the severity of machine translation's representational harm is a 4, 29.4\% said 3, 17.6\% said 5, 5.9\% said 1, and 0\% said 2. \\

\noindent \textbf{Q: } Can it cause allocational harm? Please provide an example(s) of use cases and the resultant allocational harm. (text field) \\

\noindent \textbf{Q: } What’s the severity of the allocational harm? (1-5 scale) \\

\noindent Respondents argued that if machine translation is used in medical or legal contexts, a translation that automatically applies incorrectly gendered terms can result in incorrect care or invalidation. An example provided was ``a nonbinary AFAB person might not be asked about their pregnancy status when being prescribed a new medication if a cross-lingual messaging system assigned `male' terms to them''. Furthermore, non-binary persons might be ``denied a visa or convicted of a crime due to mistranslation of evidence''. Another very real consequence of machine translation systems misgendering that respondents brought up is that they can deny non-binary persons ``gender euphoria'' (i.e. the joy of having one's gender affirmed) and cause psychological harm.

Overall, on a scale of 1-5, where 1 indicates ``No impact on lives'', 3 indicates ``Noticeably negatively affects lives'', and 5 indicates ``Significantly hinders lives'', 33.3\% of respondents said that the severity of machine translation's allocational harm is a 5, 20\% said 2, 20\% said 3, 20\% said 4, and 6.7\% said 1. \\

\noindent \textbf{Q: } Rank the representational harms caused by the aforementioned tasks by severity for the worst realistic use case. \\

\begin{table*}[]
    \centering
    \begin{tabular}{|c|c|c|c|}
    \hline
    \diagbox{\textbf{Severity Ranking}}{\textbf{Task}} & \textbf{NER} & \textbf{Coreference Resolution} & \textbf{Machine Translation} \\
        \hline
         Lowest & 52.6\% & 21.1\% & 21.1\% \\
         In-Between & 21.1\% & 42.1\% & 31.6\% \\
         Highest & 21.1\% & 31.6\% & 42.1\% \\
         \hline
    \end{tabular}
    \caption{Language Task Rankings by Severity of  Representational Harm}
    \label{tab:repharmsranking}
\end{table*}

\noindent Results may be found in Table \ref{tab:repharmsranking}. \\

\noindent \textbf{Q: } Rank the allocational harms caused by the aforementioned tasks by severity for the worst realistic use case. \\

\begin{table*}[]
    \centering
    \begin{tabular}{|c|c|c|c|}
    \hline
    \diagbox{\textbf{Severity Ranking}}{\textbf{Task}} & \textbf{NER} & \textbf{Coreference Resolution} & \textbf{Machine Translation} \\
        \hline
         Lowest & 36.8\% & 26.3\% & 26.3\% \\
         In-Between & 21.1\% & 42.1\% & 26.3\% \\
         Highest & 31.6\% & 21.1\% & 36.8\% \\
         \hline
    \end{tabular}
    \caption{Language Task Rankings by Severity of  Allocational Harm}
    \label{tab:allocharmsranking}
\end{table*}

\noindent Results may be found in Table \ref{tab:allocharmsranking}.

\subsection{Broader Concerns with Language Models}
\label{sec:App broader concerns}

This segment was purposely kept less specific to understand the harms in different domains (healthcare, social media, etc.) and their origins, as perceived by different non-binary individuals. \\

\noindent \textbf{Q: } Can you see/think of domains (e.g. healthcare, social media, public administration, high-tech devices, etc.) to which language models can/could be *applied* in a way that produces undesirable outcomes for non-binary individuals? If so, please list such domains below. (text field) \\

\noindent \textbf{Q: } For each domain you listed above, please provide an example(s) of harmful applications and use cases and evaluate the severity of the resultant harms. \\

\noindent \textbf{Social Media} \\

\noindent LGBTQ+ social media content is automatically flagged at higher rates. Ironically, language models can fail to identify hateful language targeted at nonbinary people. Furthermore, if social media sites attempt to infer gender from name or other characteristics, this can lead to incorrect pronouns for non-binary individuals. Additionally, ``language models applied in a way that links entities across contexts are likely to out and/or deadname people, which could harm trans and nonbinary people''. Moreover, social media identity verification could incorrectly interpret non-binary identities as fake or non-human. \\

\noindent \textbf{Productivity Technologies} \\

\noindent Autocomplete could suggest ``only binary pronouns, or make predictions that align with gender stereotypes''. \\

\noindent \textbf{Healthcare} \\

\noindent Respondents said that ``healthcare requires engaging with gender history as well as identity'', which language models are not capable of doing, and ``even humans intending to do well and using the best terms they know often struggle with the limitations of our language for nonbinary people and their bodies''. Language models could further ``misgender patients''. Additionally, language models could ``deny insurance claims, e.g. based on a `mismatch' between diagnosis and gender/pronouns''. \\

\noindent \textbf{Policing} \\

\noindent Respondents said that ``any system which incorrectly handles singular `they' might result in communications being flagged as false, self-contradictory, or incomplete''. \\

\noindent \textbf{Marketing and Customer Service} \\

\noindent Language models could enable ``predatory or adversarial advertising'' for non-binary persons. \\

\noindent \textbf{Hiring} \\

\noindent Respondents explained that ``a system which incorrectly handles singular `they' might result in non-binary people's achievements being misattributed to group work or to organizations they worked for''. \\

\noindent \textbf{Finance} \\

\noindent Finance-related identity verification could incorrectly interpret non-binary identities as fake or non-human. \\

\noindent \textbf{Government-Administrated Services} \\

\noindent Government services could misgender non-binary persons or reject their applications based on language analysis. \\

\noindent \textbf{Education} \\

\noindent Language models employed in automated educational/grading tools could ``automatically mark things wrong/`ungrammatical' for use of non-standard language, singular `they', neopronouns, and other `new' un- or creatively gendered words''. \\


\noindent \textbf{Q: } Can you see/think of applications of language models that can/could exacerbate non-binary erasure? If so, please list such applications below. (text field) \\

\noindent \textbf{Q: }  For each application you listed above, please provide an example(s) of harmful use cases and evaluate the severity of the resultant harms using the 1-5 scale below. (text field) \\

\noindent Automated summarization (e.g. used in Google’s information boxes) could erase non-binary persons. For example, non-binary people are ``more likely to be tagged as non-human and thus less likely to have their achievements accurately summarized, which makes them make invisible''.

Moreover, current language models cannot generate text with nonbinary people or language (e.g. "it never generates sentences with `they/them' pronouns or `ze/hir' pronouns or a sentence like `She is nonbinary', but it regularly generates examples with `he/him' and `she/her' pronouns and sentences like `He is a man'); this decidedly contributes to nonbinary erasure.

Screen readers and speech-to-text services that cannot handle neopronouns may also erase non-binary individuals. Similarly, ``neopronouns are almost always listed as `wrong''' by spelling and grammar checkers. 

Machine translation is particularly prone to erasing non-binary gender because ``nonbinary people often create new ways of using language out of necessity, and these usages are rare/new enough to not be reflected in machine translation''.

One respondent said that ``any model that attempts to classify gender'' contributes to nonbinary erasure because ```nonbinary' is not a single entity or identity type''; further, ``treating `nonbinary' as a distinct third gender or as some `ambiguous' category'' is also erasing. \\

\noindent \textbf{Cisgender Privilege} is the unearned benefits you receive when your gender identity matches your sex assigned at birth. \\

\noindent \textbf{Q: } Can you see/think of applications of language models that can/could exacerbate transphobia or denial of cisgender privilege? If so, please list such applications below. (text field) \\

\noindent \textbf{Q: } For each application you listed above, please provide an example(s) of harmful use cases and evaluate the severity of the resultant harms. (text field) \\

\noindent Respondents said that language models can exacerbate transphobia ``by incorrectly flagging non-toxic content from trans persons as toxic at higher rates, or by not recognizing transphobic comments as toxic''. 

Furthermore, ``any system that attempts to ascertain gender or pronouns from a name or other attributes'' can enable cisgender privilege.  A respondent explained that ``if a model misgenders someone because it accounts for a history of them having another name, or does not allow for flexibility in gender signifiers to change over time, it reinforces the idea that gender is or should be immutable, that individuals have a `true' gender and then `change it,' that gender can only `change' once if it happens at all, and that there is some clear point of demarcation where you `transition' and only ever in one direction (binary transition)''; furthermore, ``any corrections to those assumptions in the model would necessarily be post-hoc, marking oneself as `other' for not fitting into the binary construction'' of gender.

Additionally, there are dangerous applications, like bots on social media, that systematically harass non-binary people.

Language models can also be used to empower the enforcement mechanisms of transphobic policies. This could occur in ``visual/textual systems for things like passport verification or other legal processing like getting driver's licenses, applying for loans, attempting to obtain citizenship''.

While developing and testing language model-based systems, developers may find that the language nonbinary persons have created for themselves is not compatible with their systems. Hence, developers may ``blame nonbinary people'' for the difficulty associated with including them and ``decide that the systems will only serve binary-aligned people''. However, ``this both increases cis/binary privilege (by making those systems inaccessible to nonbinary people) and increases transphobia (by creating or strengthening feelings of resentment towards people who do not fit conveniently into the gender binary)''. \\

\noindent \textbf{Q: } What are the top reasons that there exist limited data concerning non-binary individuals in Natural Language Processing (NLP) in your opinion? \\

\noindent Most respondents cited a lack of diversity in developer/research teams. They said there exists ``limited trans/non-binary representation so knowledge gaps exist'', and many developers and researchers have a ``lack of knowledge about nonbinary identities, transness, queerness, etc.'' Further, developer/research teams tend to ``want to simplify variables and systems after and in spite of learning about the complexity'' of gender identity, and may not consider non-binary persons ``important enough to change their systems for''.

Respondents also discussed the sources of training data. They explained that ``most training data pulls from large scale internet sources and ends up representing hegemonic viewpoints''. Additionally, ``lots of our models are built on Wikipedia which has few non-binary people, Reddit which is hardly a queer utopia, and books written when gender non-conforming content did not get published much''. Moreover, non-binary data may be ``discarded as `outliers''' and ``not sampled in training data''. Data annotators may also ``not recognize non-binary identities'', ``non-binary identities may not be possible labels'',  and/or annotators ``may not be paying attention to non-binary identities due to minimal wages, lack of situational context, and lack of epistemic uncertainty in the models''.

Other major reasons included ``historic erasure, active discrimination, invisibility of some non-binary identities, small non-binary population''. \\

\noindent From \citet{barocas-hardt-narayanan}: \\
\noindent \textbf{Skewed sampling:} model-influenced, positive feedback loops in data collection \\
\noindent \textbf{Tainted examples:} historically-biased training data \\
\noindent \textbf{Limited features:} missing or erroneous features for training examples \\
\noindent \textbf{Sample size disparities:} insufficient training examples for minority classes \\
\noindent \textbf{Proxies:} correlated features leak undesirable information \\

\noindent \textbf{Q: } Score the following barriers to better including non-binary individuals in language models using this 1-5 scale: 1 (Easy solutions that could be deployed immediately),
3 (Could eventually achieve solutions),
5 (Impossible to remedy) \\

\begin{table*}[]
    \centering
    \begin{tabular}{|c|c|c|c|c|c|}
    \hline
    \diagbox{\textbf{Source of Bias}}{\textbf{Feasibility Ranking}} & \textbf{1} & \textbf{2} & \textbf{3} & \textbf{4} & \textbf{5} \\
    \hline
    Skewed Sampling & 0\% & 36.8\% & 47.4\% & 5.3\% & 0\% \\
    Tainted Examples & 5.3\% & 26.3\% & 47.4\% & 5.3\% & 5.3\% \\
    Limited Features & 10.5\% & 21.1\% & 21.1\% & 47.4\% & 10.5\% \\
    Sample Size Disparities & 5.3\% & 36.8\% & 26.3\% & 26.3\% & 0\% \\
    Proxies & 5.3\% & 5.3\% & 31.6\% & 36.8\% & 10.5\% \\
    \hline
    \end{tabular}
    \caption{Feasibility of Mitigation Rankings for Sources of Bias}
    \label{tab:feasibility}
\end{table*}

\noindent Results may be found in Table \ref{tab:feasibility}. \\

\noindent \textbf{Q: } Can you see/think of cases where harms (representational or allocational) are compounded for non-binary individuals with particular intersecting identities? (radio buttons) \\
\noindent \textbf{Options: } Yes, No \\

\noindent 89.5\% of respondents indicated Yes, while 10.4\% indicated No. \\

\noindent \textbf{Q: } If Yes, could you give examples of such intersecting identities? (text field) \\

\noindent \textbf{Q: } For each intersecting identity, please provide an example(s) of harmful use of language models and evaluate the severity of the resultant harm. (text field) \\

\noindent Issues with coreference resolution could be compounded for non-binary persons with non-Western names. In addition, machine translation can fail for someone when translating ``from a culture where their nonbinary identity does not fit neatly into the nonbinary boxes we've devised in English''. Non-binary racial minorities in Western societies will also be misrepresented and underrepresented in data samples. And, non-white non-binary individuals are more susceptible to misgendering and related harms in policing that employs language models.

Furthermore, medical harms can be worsened for non-binary persons who already have limited access to healthcare due to other aspects of their identity, like race, immigration status, fluency in English, etc. Moreover, non-binary persons with limited fluency in a language who have more interactions with machine translation systems are more likely to regularly incur the aforementioned representational and allocational harms posed by the systems. 

Additionally, ``some neurodivergent people refer to themselves with traditionally dehumanizing language, which could compound the issue of models not recognizing their identities as real and human'' if they're also non-binary. Further, non-binary persons with certain disabilities who rely on language model-based speech-to-text services may not have their pronouns recognized.

Other examples of intersecting identities included: class, body size, religious affiliation, nationality, sexual or romantic orientation, age, and education level. \\

\section{Dataset Skews}

\label{App: Dataset skews}
 Usage of non-binary pronouns in text is not always meaningful with respect to gender, as seen in Table \ref{tab2}.

\begin{table*}[htbp]
\centering
\caption{Example sentences containing nonbinary pronouns}
\small
\begin{tabular}{|c|p{10cm}|}\hline
\textbf{Pronoun} & \textbf{Sentence}\\\hline
Ey& \textit{``The difference in the alphabets comes only in the Faroese diphthongs (ei being 26, ey 356, oy 24...).''}\\\hline
Em& \textit{Approximating the em dash with two or three hyphens.}\\\hline
Xem& \textit{```Em đi xem hoi trang ram''', establishing her icon for Vietnamese women as well as earning the title of the ```Queen of Folk'''}\\\hline
Ze& \textit{``He taught himself to write with his left hand and described his experiences before, during, and after the accident in a deeply moving journal, later published under the title `Pogodzic sie ze swiatem' (`To Come to Terms with the World').'', }\\\hline
Zir& \textit{``The largest operation in the Struma Valley was the capture by 28th Division of Karajakoi Bala, Karajakoi Zir and Yenikoi in October 1916.''}\\\hline
\end{tabular}
\normalsize
\label{tab2}
\end{table*}

Further, the distribution of different pronouns is also not equal across genders. Overall, using the Python library $wordfreq$ \url{https://pypi.org/project/wordfreq/} which samples over diverse data to give an approximate usage of different words in all of the text curated from the web, we observe how vastly different the frequencies of different gendered words are per billion words in English \cite{robyn_speer_2018_1443582}. While `he' and `she' occur 0.49\% and 0.316\% per billion words respectively, the percent for `xe' and `ze' is only 0.0005\% and 0.0011\% respectively. We list these percentages for a larger set of gendered words in Appendix \ref{tbl: Dataset skews} to highlight this disparity.

\begin{table*}[]
\caption{Per Billion Word Frequency in the English Language}
\small
\label{tab:billion}
	\begin{center}
    \begin{tabular}{|l|r|}
    \hline
     \textbf{Word}        &   \textbf{Frequency (\%)} \\
    \hline
     he          &     0.49    \\
     his         &     0.324   \\
     they        &     0.316   \\
     she         &     0.182   \\
     them        &     0.155   \\
     man         &     0.0661  \\
     girl        &     0.024   \\
     woman       &     0.0224  \\
     himself     &     0.0178  \\
     boy         &     0.0148  \\
     female      &     0.01    \\
     male        &     0.00776 \\
     herself     &     0.00603 \\
     two-spirit  &     0.00588 \\
     em          &     0.00372 \\
     hers        &     0.00093 \\
     transgender &     0.00081 \\
     queer       &     0.00057 \\
     ey          &     0.00019 \\
     ze          &     0.00012 \\
     xe          &     5e-05   \\
     nonbinary   &     2e-05   \\
     cisgender   &     2e-05   \\
     genderqueer &     1e-05   \\
     zem         &     1e-05   \\
     genderfluid &     1e-05   \\
     xem         &     0       \\
     zey         &     0       \\
     zir         &     0       \\
     bigender    &     0       \\
     xir         &     0       \\
     cisman      &     0       \\
     ciswoman    &     0       \\
     xey         &     0       \\
    \hline
    \end{tabular}
	\end{center}
	\label{tbl: Dataset skews}
\end{table*}

\subsection{Representation Skews}
\label{app: representational skews}

Glove was trained on English Wikipedia \footnote{\url{https://dumps.wikimedia.org/}} articles with a window size of 15, a dimension of 50 for each word, and a minimum word frequency of 5. Skews as seen in GloVe representations are seen here with respect to nearest neighbors in Table \ref{tbl: nearest nbrs possessive} and often even with derogatory associations reflecting social biases (Table \ref{nbrs1}).

\begin{table*}[]
\centering
\caption{Five Nearest neighbors  for binary and non-binary possessive pronouns}
\small
\begin{tabular}{|c|c|}\hline
\textbf{Pronoun} & \textbf{Top 5 Neighbors}\\\hline
His& \textit{'he', 'him', 'who', 'after', 'himself'}\\\hline
Hers& \textit{'somehow', 'herself', 'thinks', 'someone', 'feels'}\\\hline
Theirs& \textit{'weren', 'tempted', 'couldn', 'gotten', 'willingly'}\\\hline
Xers& \textit{''yogad', 'doswelliids', 'hlx', 'cannibalize', 'probactrosaurus'}\\\hline
Zers & \textit{'ditti', 'bocook', 'kurikkal', 'felimy', 'hifter'}\\\hline
Eirs & \textit{'cheor', 'yha', 'mnetha', 'scalier', 'paynet'}\\\hline
\end{tabular}
\label{tbl: nearest nbrs possessive}
\normalsize
\end{table*}

\begin{table*}[]
\centering
\caption{Ten Nearest neighbors of non-binary terms highlighting derogatory Terms}
\label{App: derogatory}
\small
\begin{tabular}{|l|l|}
\hline
 \textbf{Term}        & \textbf{10 Nearest Neighbors}                                                                                                  \\
\hline
 agender     & bigender, genderfluid, genderqueer, tosin, cisgender, nonbinary, laia, muhafazat, \textbf{negrito}, farmgirl                   \\
 bigender    & pangender, agender, genderfluid, overcontact, pnong, genderqueer, nonbinary, eczemas, gegs                 \\
 queer       & lesbian, lgbtq, feminism, lgbt, lesbians, feminist, racism, sexuality, stereotypes, gay                               \\
 nonbinary   & genderqueer, \textbf{transsexual}, cisgender, \textbf{transsexuals}, bigender, genderfluid, chorti, referents, pansexual, hitchhikers \\
 transgender & lesbian, lgbt, lgbtq, bisexual, intersex, gender, \textbf{transsexual}, lesbians, heterosexual, discrimination                 \\
 genderfluid & agender, bigender, genderqueer, transwoman, nonbinary, pansexual, montserratian, \textbf{negrito}, supercouple, \textbf{fasiq}          \\
 genderqueer & pansexual, nonbinary, lgbtqia, \textbf{transsexual}, genderfluid, agender, bisexuality, bigender, diasporic, multiracial       \\
\hline
\end{tabular}
\label{nbrs1}
\end{table*}

\paragraph{Biases With Respect to Occupations}
Binary gender and their stereotypically associated occupations is a bias widely discussed. We see in Table \ref{tbl: occupations} that is not very relevant for non-binary-gendered persons and the biases faced by them.



\begin{table*}[h]
\small
        \begin{adjustbox}{width=\textwidth,totalheight=\textheight,keepaspectratio}
		\begin{tabular}{|c|c|c|c||c|c|c|c|c|c|c|} \hline
			occupation & man & woman & transman & transwoman & cisgender & transgender & nonbinary & genderqueer & genderfluid & bigender \\
			\hline
			doctor & 0.809 & 0.791 & -0.062 & -0.088 & 0.094 & 0.388 & 0.037 & 0.022 & 0.069 & -0.107 \\
			engineer & 0.551 & 0.409 & -0.152 & -0.271 & -0.227 & 0.043 & -0.243 & -0.176 & -0.084 & -0.298 \\
			nurse & 0.616 & 0.746 & -0.095 & 0.050 & 0.206 & 0.527 & 0.129 & 0.083 & 0.182 & 0.022 \\
			stylist & 0.382 & 0.455 & 0.018 & 0.062 & 0.117 & 0.318 & 0.015 & 0.126 & 0.207 & -0.017 \\
			\hline
		\end{tabular}
		\end{adjustbox}
		\normalsize
	\caption{Cosine similarity between occupations and words}
	\label{tbl: occupations}
\end{table*}

\begin{table*}[]
\label{App: adjectives}
\small
	\begin{center}
		\begin{tabular}{|l|l|} \hline
			\textbf{Set} & \textbf{Adjectives} \\
			\hline
			positive & \textit{smart, wise, able, bright, capable, ambitious, calm, attractive, great, good, caring, loving, adventurous} \\
			negative & \textit{dumb, arrogant, careless, cruel, coward, boring, lame, incapable, rude, selfish, dishonest, lazy, unkind} \\
			\hline
		\end{tabular}
	\end{center}
	\caption{Positive and Negative Adjective Sets}
\end{table*}

\begin{table*}[]
	\begin{center}
		\begin{tabular}{|c|c|}
		    \hline
			\textbf{Proxy} & \textbf{Average} \\
			\hline
			man & 0.462 \\
			woman & 0.494 \\
			transman & -0.043 \\
			transwoman & 0.088 \\
			cisgender & 0.101 \\
			transgender & 0.228 \\
			nonbinary & 0.025 \\
			genderqueer & 0.054 \\
			genderfluid & 0.126 \\
			bigender & -0.052 \\
			\hline
		\end{tabular}
	\end{center}
	\caption{Average similarity between occupations and words}
\end{table*}

\noindent \paragraph{\textbf{Subspace analyses}}

\begin{figure}
\label{subspace}
\vskip 0.2in
\begin{center}
\includegraphics[width=\columnwidth]{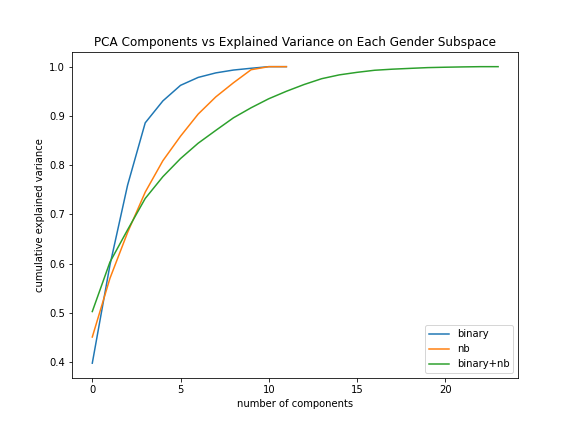}
\caption{PCA Components for each Gender Subspace}
\end{center}
\vskip -0.2in
\end{figure}

Capturing a gender subspace has been useful in techniques of bias analysis and techniques in subsequent debiasing in binary gender~\cite{dev2019attenuating}, especially in context-free or static representations like GloVe or word2vec. These methods postulate expanding this to nonbinary gender by determining a general subspace for gender which captures both binary and non-binary genders. We test if we can approach capturing the all-gender subspace by extending one such general subspace capturing method~\cite{bolukbasi2016man} - principal component analysis (PCA) - on 3 groups of words: 

\begin{enumerate}
    \item \textbf{Binary set}: 'he', 'she', 'man', 'woman', 'hers', 'his', 'herself', 'himself', 'girl', 'boy', 'female', 'male'
    \item \textbf{Nonbinary set}: 'they', 'them', 'xe', 'ze', 'xir', 'zir', 'xey', 'zey', 'xem', 'zem', 'ey', 'em'
    \item \textbf{Binary + Non-Binary set}
\end{enumerate}

If we truly captured the gender subspace, we could safely assume that the difference between the binary subspace and the all-gender subspace, along with the non-binary subspace and the all-gender subspace, is somewhat negligible. We make the following observations leveraging the cosine distance, defined as $1 - c$, where $c$ is the cosine similarity between two vectors. We observe, opposite to what we expected, that the distance was quite different in these respective pairs. Between the binary and all-gender subspace was a cosine distance of 1.48, while the distance between the non-binary and all-gender subspace was larger, at 1.93.
This tells us that the binary subspace is much less dissimilar than the nonbinary subspace with respect to the all-gender subspace, i.e., extending the approach of subspace capture to all genders would result in a subspace more dominantly aligned with binary gender than non-binary gender. Further, due to the poor representation of non-binary pronouns, the subspace is likely representing the difference in frequency of terms rather than the concept of gender as a whole. Due to weaker alignment with the non-binary gender, any tasks performed using this new `gender' subspace would not be very effective or applicable to non-binary genders, thus indicating towards further skews and harm. 

\subsection{Words for WEAT and Similarity Tests}
\label{app: WEAT}

WEAT tests require groups of words. We list herein Tables \ref{App: binary_words_def}, \ref{app pleasant/unpleasant}, \ref{tbl: weat scores} the pronouns and words used as well as the pleasant and unpleasant words we compared them against.

\begin{table*}[]

\small
	\begin{center}
		\begin{tabular}{|l|l|} \hline
			\textbf{Set} & \textbf{Words} \\
			\hline
			binary pronouns & \textit{he, him, his, she, her, hers}\\
			binary words & \textit{man, woman, herself, himself, girl, boy, female, male, cisman*, ciswoman*}\\
			binary all & \textit{binary pronouns + binary words}\\
			
			nonbinary pronouns & \textit{zey, ey, em, them, xir, they, zem, ze, their, zir, zers, eirs, xey, xers, xe, xem} \\
			nonbinary words & \textit{transgender, queer, nonbinary, genderqueer, genderfluid, bigender, two-spirit}\\
			nonbinary all & \textit{nonbinary pronouns + nonbinary words}\\
			\hline
		\end{tabular}
	\end{center}
	\caption{Word set definitions for binary and non-binary concepts}
	\label{App: binary_words_def}
\end{table*}

\begin{table*}[]
\label{App: plesant_words}
\small
	\begin{center}
		\begin{tabular}{|l|l|} \hline
			\textbf{Set} & \textbf{Words} \\
			\hline
			pleasant & \textit{joy, love, peace, wonderful, pleasure, friend, laughter, happy} \\
			unpleasant & \textit{agony, terrible, horrible, nasty, evil, war, awful, failure} \\
			\hline
		\end{tabular}
	\end{center}
	\caption{Set of unpleasant and pleasant words}
	\label{app pleasant/unpleasant}
\end{table*}

\begin{table*}[]
\small
	\begin{center}
		\begin{tabular}{|c|c|c|} \hline
			\textbf{Words} & \textbf{Average} & \textbf{Absolute Average} \\
			\hline
			he & 0.509 & 0.509 \\
			him & 0.465 & 0.465 \\
			his & 0.498 & 0.498 \\
			she & 0.495 & 0.495 \\
			her & 0.473 & 0.473 \\
			xir & -0.197 & 0.207 \\
			they & 0.395 & 0.395 \\
			xey & -0.007 & 0.094 \\
			them & 0.389 & 0.390 \\
			ey & 0.086 & 0.111 \\
			zey & -0.056 & 0.108 \\
			xe & -0.054 & 0.111 \\
			their & 0.378 & 0.379 \\
			xers & -0.088 & 0.105 \\
			em & 0.185 & 0.209 \\
			zir & -0.035 & 0.092 \\
			zem & -0.068 & 0.091 \\
			eirs & -0.158 & 0.185 \\
			zers & -0.104 & 0.116 \\
			ze & 0.123 & 0.143 \\
			xem & -0.169 & 0.180 \\\hline
		\end{tabular}
	\end{center}
	\label{app: cos occ pronouns}
	\caption{Average cosine similarity between occupations and pronouns}
\end{table*}

\begin{table*}[]
\small
	\begin{center}
		\begin{tabular}{|c|c|c|}\hline
			\textbf{Words} & \textbf{Average} & \textbf{Absolute Average} \\
			\hline
			man & 0.469 & 0.469 \\
			woman & 0.473 & 0.473 \\
			herself & 0.400 & 0.400 \\
			himself & 0.483 & 0.483 \\
			girl & 0.421 & 0.421 \\
			boy & 0.457 & 0.457 \\
			female & 0.393 & 0.394 \\
			male & 0.350 & 0.353 \\
			transman & -0.052 & 0.098 \\
			transwoman & 0.023 & 0.125 \\
			transgender & 0.262 & 0.262 \\
			queer & 0.184 & 0.192 \\
			nonbinary & 0.010 & 0.088 \\
			genderqueer & 0.048 & 0.099 \\
			genderfluid & 0.079 & 0.113 \\
			bigender & -0.085 & 0.118 \\
			\hline
		\end{tabular}
	\end{center}
	\label{app: cos occ words}
	\caption{Average cosine similarity between occupations and words}
\end{table*}

\begin{table}[t]
\centering
\small
\begin{tabular}{|p{5cm}|c|}
\hline
\textbf{Sets} & \textbf{Weat Score} \\\hline
Random Vectors & -0.02 \\\hline
Binary Pronouns vs. Non-Binary Pronouns & 0.2  \\\hline
Binary Words vs. Non-Binary Proxies & 0.718 \\\hline
Binary Pronouns + Words vs. Non-Binary Pronouns + Proxies & 0.916  \\\hline

\end{tabular}
\caption{WEAT Scores (vs. pleasant and unpleasant attributes)}

\label{tbl: weat scores}
\normalsize
\end{table}

\subsection{BERT experiments}

We create a balanced, labeled dataset containing sentences using either $they(s)$ or $they(p)$. The text spans for $they(p)$ are chosen randomly from Wikipedia containing pairs of sentences such that the word $they$ appears in the second sentence (with no other pronoun present) and the previous sentence has a mention of two or more persons (determined by NER). This ensures that the word $they$ in this case was used in a plural sense. For the samples with singular usage of $they$, since Wikipedia does not have a large number of such sentences in general, we randomly sample them from the Non-Binary Wiki \footnote{\url{https://en.wikipedia.org/wiki/Category:Non-binary_gender}}. The sentences are manually annotated for further confirmation of correct usage of each pronoun.
We follow the procedure of data collection for $they(s)$ to create datasets for sentences with words $he$ and $she$ from Wikipedia. In each dataset, we replace the pronouns with the $[MASK]$ token and use BERT to predict the token's representation.
Figure \ref{fig:BERT1} demonstrates that the representations of $he$, $she$ and $they(s)$ acquired in this manner are similarly separable from the representations of $they(p)$.
\begin{figure*}[h]
  \centering
  \includegraphics[scale=0.5]{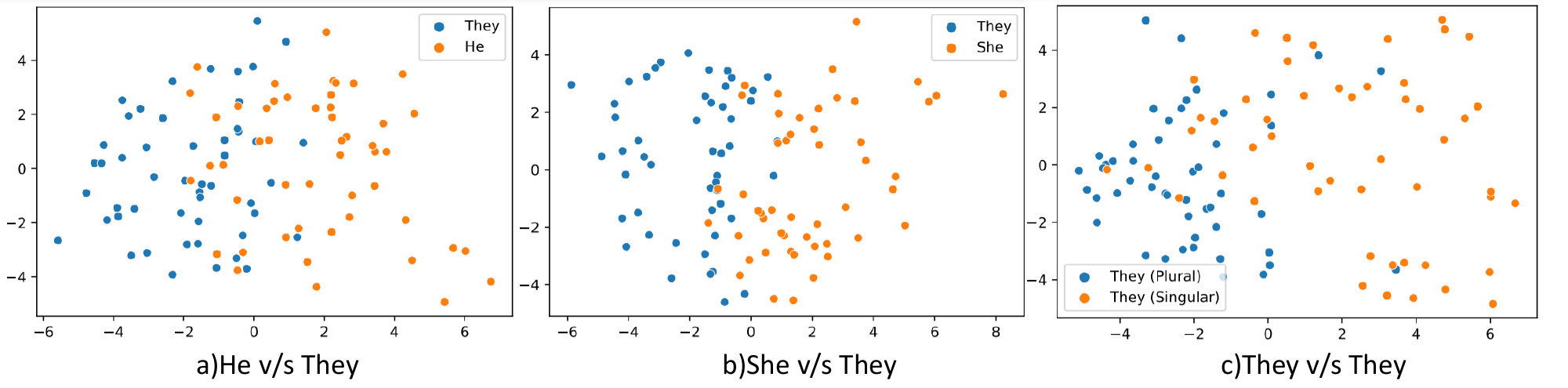}
  \caption{BERT representation analysis: a)he v/s they(p) b)she v/s they(p) c)they(s) v/s they(p)}
\label{fig:BERT1}
\end{figure*}

\label{app: bert}
\begin{table}[]
    \centering
    \small
    \begin{tabular}{|c|c|}
    \hline
        \textbf{Classifier} & \textbf{Accuracy} \\
        \hline
         $C_1$ & 67.7\%  \\
         $C_2$ & 83.3\% \\
         $C_3$ & 83.1\%  \\ \hline
    \end{tabular}
    \normalsize
    \caption{The performance of BERT classifier}
    \label{tab:bert2}
\end{table}

\begin{figure}[h]
  \centering
  \includegraphics[scale=0.5]{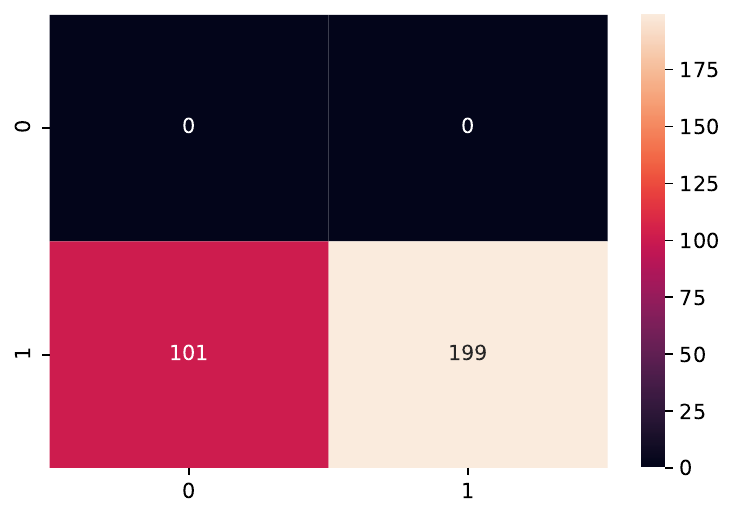}
  \caption{Confusion matrix of Classifier C1}
\label{fig:BERT1}
\end{figure}

\begin{figure}[h]
  \centering
  \includegraphics[scale=0.5]{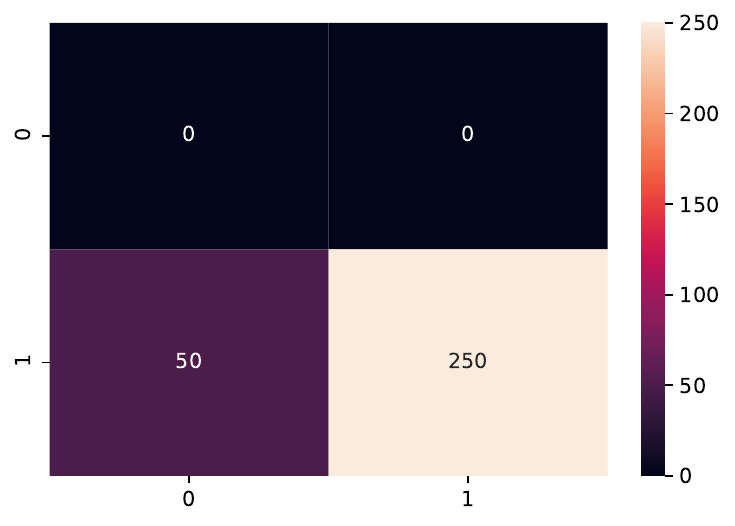}
  \caption{Confusion matrix of Classifier C2}
\label{fig:BERT1}
\end{figure}

\begin{figure}[h]
  \centering
  \includegraphics[scale=0.5]{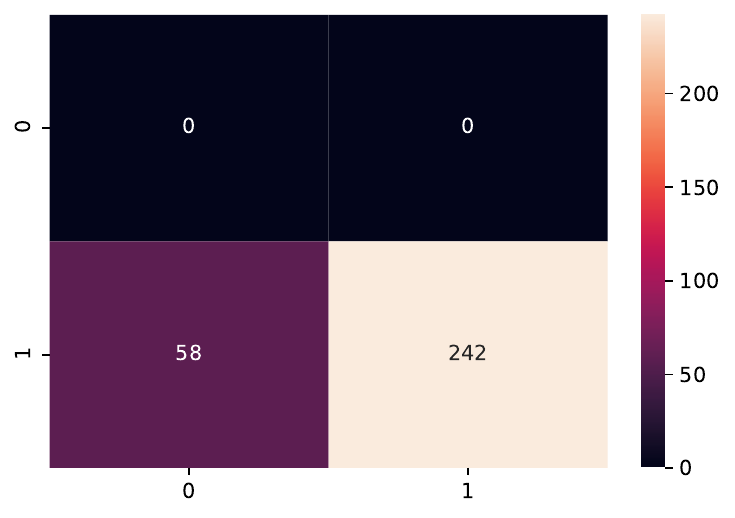}
  \caption{Confusion matrix of Classifier C3}
\label{fig:BERT1}
\end{figure}

\paragraph{BERT templates for Misgendering}
We used templates of sentences to evaluate if misgendering is propagated by BERT. 
The sentence templates used are:

\noindent [Name] went to the hospital for [pronoun] appointment. [MASK] was feeling sick.\\
\noindent [Name] went to the salon for [pronoun] appointment. [MASK] got a haircut.\\
\noindent [Name] went to the university for [pronoun] classes. [MASK] had a morning class.\\
\noindent [Name] went to the restaurant near [pronoun] house. [MASK] bought lunch.\\
\noindent [Name] drove to the hospital for [pronoun] appointment. [MASK] was feeling sick.\\
\noindent [Name] drove to the salon for [pronoun] appointment. [MASK] got a haircut.\\
\noindent [Name] drove to the university for [pronoun] classes. [MASK] had a morning class.\\
\noindent [Name] drove to the restaurant near [pronoun] house. [MASK] bought lunch.\\
\noindent [Name] walked to the hospital for [pronoun] appointment. [MASK] was feeling sick.\\
\noindent [Name] walked to the salon for [pronoun] appointment. [MASK] got a haircut.\\
\noindent [Name] drove to the university for [pronoun] classes. [MASK] had a morning class.\\
\noindent [Name] fed [pronoun] dog. [MASK] had to leave for work.\\
\noindent [Name] met [pronoun] friend at the cafe. [MASK] ordered a coffee.\\
\noindent [Name] attached a file to [pronoun] email. [MASK] sent the email.\\
\noindent [Name] realized [pronoun] left [pronoun] keys at home. [MASK] ran back to get the keys.\\
\noindent [Name] found [pronoun] drivers license on the pavement. [MASK] picked it up.\\
\noindent [Name] checks [pronoun] phone constantly. [MASK] is expecting an important email.\\
\noindent [Name] said that [pronoun] child was just born. [MASK] is excited for the future.\\
\noindent [Name] is in a rush to attend [pronoun] lecture. [MASK] eats lunch quickly.\\
\noindent [Name] enjoys riding [pronoun] bike. [MASK] is able to get anywhere.\\

We vary the name (over 900 names) and pronouns as described in the paper in Section \ref{sec: representation}.


\end{document}